\documentclass[10pt,twocolumn,letterpaper]{article}

\usepackage{iccv}
\usepackage{times}
\usepackage{epsfig}
\usepackage{graphicx}
\usepackage{amsmath}
\usepackage{amssymb}

\usepackage{booktabs}
\usepackage[ruled,vlined]{algorithm2e}
\usepackage{mathrsfs}
\DeclareMathOperator*{\argmax}{arg\,max}

\makeatletter
\newcommand*\bigcdot{\mathpalette\bigcdot@{.5}}
\newcommand*\bigcdot@[2]{\mathbin{\vcenter{\hbox{\scalebox{#2}{$\m@th#1\bullet$}}}}}
\makeatother

\usepackage[pagebackref=true,breaklinks=true,letterpaper=true,colorlinks,bookmarks=false]{hyperref}

\newcommand*{\affaddr}[1]{#1} 

\iccvfinalcopy 

\def\iccvPaperID{8919} 

\ificcvfinal\fi

\begin{document}

\title{Uncertainty-Aware Model Adaptation for Unsupervised\\ Cross-Domain Object Detection}

\author{Minjie Cai,~~Minyi Luo,~~Xionghu Zhong,~ and~ Hao Chen\\
\affaddr{College of Computer Science and Electronic Engineering, Hunan University}\\
{\tt\small \{caiminjie, anaser, xzhong, haochen\}@hnu.edu.cn}
}

\maketitle
\ificcvfinal\thispagestyle{empty}\fi

\begin{abstract}
   This work tackles the unsupervised cross-domain object detection problem which aims to generalize a pre-trained object detector to a new target domain without labels. We propose an uncertainty-aware model adaptation method for object detection, which is based on two motivations: 1) the estimation and exploitation of model uncertainty in a new domain is critical for reliable domain adaptation; and 2) the joint alignment of distributions for inputs (feature alignment) and outputs (model self-training) is needed. To this end, we compose a Bayesian CNN-based framework for uncertainty estimation in object detection, and propose an algorithm for generation of uncertainty-aware pseudo-labels. We also devise a scheme for joint feature alignment and self-training of the object detection model with pseudo-labels. Experiments on multiple cross-domain object detection benchmarks show that our proposed method achieves state-of-the-art performance.
\end{abstract}

\section{Introduction}
Object detection is a popular problem in computer vision, and has wide applications in various domains such as video surveillance and autonomous driving \etal. With large amount of annotated data and advanced deep neural networks such as Faster R-CNN \cite{ren2016faster} and YOLO \cite{yolov3}, the performance of object detection has been greatly promoted. However, it still remains challenging to deploy a pre-trained object detector in new unseen domains due to the distribution discrepancy of different domains, which is common in practical applications.

Recently, many research efforts have been devoted to (unsupervised) cross-domain object detection~\cite{chen2018domain,li2020deep}, which aims to generalize object detectors trained from labeled source domain data to unlabeled target domain data. Existing approaches have attempted to align the marginal distribution of the inputs from two domains by learning domain-invariant features through adversarial learning~\cite{chen2018domain,saito2019strong-weak, xu2020categorical-regulaztion, xu2020graph-induced} . However, this does not guarantee successful domain adaptation since the joint distribution of inputs and outputs may not be aligned. As a result, objects in the target domain may be detected as incorrect categories in the source domain even though the image features of two domains are well aligned. 

In order to align the outputs as well as the inputs, an alternate strategy of domain adaptation is \textit{self-training} which treats the predictions of a pre-trained model on the target domain data as pseudo-labels for re-training the model \cite{khodabandeh2019robust}. However, there are challenges for self-training-based approaches. One big challenge is the negative influence of noisy pseudo-labels during model re-training. It is possible that wrongly detected instances are considered as labels which increases the difficulty of self-training or even deteriorate the model. Moreover, the distribution discrepancy between two domains would further increase the risk of assigning wrong pseudo-labels for self-training.

Following the strategy of self-training, we identify two important ways for addressing the above challenges. First, since the selection of ``confident'' pseudo-labels is critical for successful self-training and model uncertainty measures how confident a model is with its prediction, the estimation and exploitation of model uncertainty in the unseen target domain is desired.  Second, since data discrepancy between the source and target domains affects the prediction and hence the generation of pseudo-labels in the target domain, feature alignment and self-training need be jointly considered for domain adaptation. 

In this paper, we propose a novel uncertainty-aware model adaptation method for object detection by leveraging model uncertainty in joint self-training and feature alignment. We compose a Bayesian CNN-based object detection framework to provide more reliable estimation of model uncertainty from both the classification and regression branches. Then, an uncertainty-aware pseudo-label selection algorithm is proposed to select predicted object instances as pseudo-labels from the target domain. Finally, the object detection model is adapted to the target domain by joint self-training and feature alignment with uncertainty-aware pseudo-labels. For self-training, both categories and locations of pseudo-labeled object instances are used together with the labeled source domain data to re-train the model. For feature alignment, the locations of pseudo-labeled object instances are used for adversarial learning of instance-level domain discriminator. 

In summary, our main contributions are as follows:
\vspace{-0.5em}
\begin{itemize}
    \item A novel uncertainty-aware model adaptation method is proposed for cross-domain object detection.
\vspace{-0.5em}
    \item A Bayesian CNN-based framework is composed  for uncertainty estimation and pseudo-label selection in object detection.
\vspace{-0.5em}
    \item State-of-the-art performance is achieved on multiple benchmarks with the proposed method.

\end{itemize}

\section{Related work}

\subsection{Unsupervised domain adaptation}
Unsupervised domain adaptation (UDA) \cite{wang2018deep,wilson2020survey} has recently gained increasing interests for deep learning tasks, and various deep domain adaptation approaches have emerged. Discrepancy-based methods \cite{mmd_1,mmd_2,coral,slice} try to fine-tune the models with unlabeled target data to diminish the discrepancy between different domains. Reconstruction-based methods~\cite{constructure_1,constructure_2,constructure_3,bi-direction} often take a data reconstruction tool, such as image style translation \cite{cyclegan}, as an auxiliary task to ensure feature invariance. Adversarial-based methods \cite{gan_input_1,gan_input_2,gan_feature_1,gan_feature_2,gan_output} have also been proposed with the development of Generative Adversarial Networks(GANs) \cite{gan}. In this work, we tackle the unsupervised domain adaption problem for object detection.

\subsection{Cross-domain object detection}
Object detection is a classical task in computer vision task and has gained significant advancement with deep learning techniques. Recently, the study of domain adaptation for object detection has attracted increasing attention \cite{li2020deep} and various approaches have been developed, following the main routine in UDA. The adversarial-based methods \cite{chen2018domain, he2019multi-layer, zhu2019selective, saito2019strong-weak, xu2020categorical-regulaztion, zhang2021rpn} utilize adversarial domain classifiers for feature alignment to diminish the distribution discrepancy. \cite{chen2018domain} firstly consider domain adaptation in object detection with image and instance level classifiers. The reconstruction-based methods  \cite{hsu2020progressive,rodriguez2019style-consistency} utilize image style translation techniques, such as CycleGAN~\cite{cyclegan}, to diminish the gap between source and target domain. Recently, self-training based methods\cite{roychowdhury2019automatic, khodabandeh2019robust, zhao2020collaborative} have been proposed as a simpler yet effective strategy, which explores the pseudo-labels from the target domain to re-train model. 

In this work, we propose a novel uncertainty-aware method for cross-domain object detection by estimating and exploiting model uncertainty in object detection. 

\subsection{Uncertainty in deep neural networks}
The estimation of model uncertainty \cite{ovadia2019can} is critical for generalization and safety of deep neural networks (DNNs) and various approaches for quantifying uncertainty have been studied \cite{mixture-density-networks,non-baysian-1,non-baysian-2,non-baysian-3,vatiation-inference,dropout}. Probabilistic neural networks such as mixture density networks \cite{mixture-density-networks} tried to catch the inherent ambiguity in outputs. As a principled mechanism to provide reliable uncertainty estimates, DNNs in Bayesian frameworks have been studied for a long period, and a variety of approximation methods are proposed such as Laplace approximation~\cite{Laplace-ap}, variational inference~\cite{vatiation-inference, vatiation-inference_1,dropout, droput_2}, expectation propagation~\cite{expectation-propagation} and stochastic gradient MCMC~\cite{MCMC}. 
Recently, increasing attention has been paid to exploiting uncertainty for domain adaptation~\cite{han2019unsupervised,wen2019bayesian,mukherjee2020uncertainty}. However, these work mainly focused on the classification task.

In this work, we compose a Bayesian CNN-based framework for uncertainty estimation and uncertainty-aware pseudo-label selection in object detection.

\begin{figure*}
    \centering
    \includegraphics[width=0.95\linewidth]{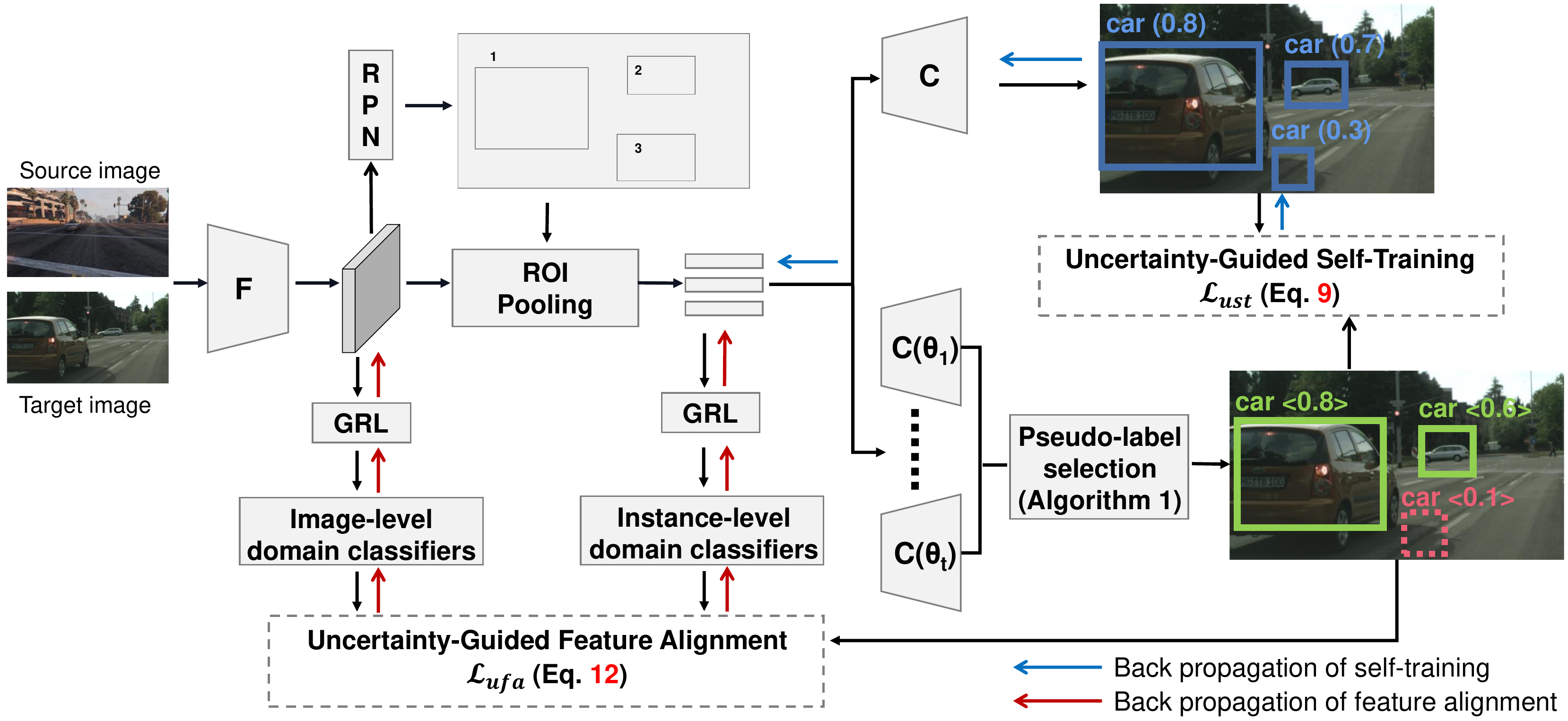}
    \caption{Overview of the proposed method. For simplicity, we only draw the data flow for input of target image. The main difference for input of source image is that ground-truth labels are used and pseudo-label selection is not needed. The green and red bounding boxes show selected and discarded pseudo-labels respectively, along with their selection scores. The blue bounding boxes show the predicted object instances with their softmax outputs. ``F'' denotes feature extractor, ``RPN'' denotes region proposal network, ``C'' denotes region-based classifier/regressor, ``GRL'' denotes gradient reversal layer.}
    \label{fig:model}
\end{figure*}

\section{Preliminaries}
\paragraph{Task definition.} Suppose we have an object detector pre-trained from the source domain data $\mathcal{D}^S = {\{ (\boldsymbol{x}_i^s,\mathcal{Y}_i^{gt}) \}}_{i=1}^{N_s}$, where $\boldsymbol{x}^s$ is a source domain image, $\mathcal{Y}^{gt}=\{\boldsymbol{y}_m\}_{m=1}^{M}$ is a set of $M$ labeled object instances, each of which involves a class label, and a location label for the object bounding box's center coordinates and its width and height. Our task is to adapt the object detector to a new target domain $\mathcal{D}^T = {\{ \boldsymbol{x}_i^t\}}_{i=1}^{N_t}$ without labels.

\paragraph{Self-training.}
Self-training was proposed for semi-supervised learning and has recently been used as a simple strategy for unsupervised domain adaptation. Self-training begins with a baseline model pre-trained on the labeled source domain data $\mathcal{D}^S$ and treats the model's predictions on the unlabeled target domain data $\mathcal{D}^T$ as pseudo-labels. The pseudo-labels are often used with $\mathcal{D}^S$ to re-train the model. The self-training loss function for a pair of source domain and target domain images is formulated as:
\begin{align}
\begin{split}
\label{st_loss}
     &\mathcal{L}_{st} = \mathcal{L}^S_{det}(\boldsymbol{x}^s, \mathcal{Y}^{gt}) + \lambda_1 \mathcal{L}^T_{det}(\boldsymbol{x}^t, \mathcal{\hat Y}^{pl}) 
\end{split}
\end{align}
where $\mathcal{L}_{det}^S$ and $\mathcal{L}_{det}^T$ denote the detection loss on the labeled source domain data and on the unlabeled target domain data with pseudo-labels $\mathcal{\hat Y}^{pl}$ respectively. $\lambda_1$ is a weight factor.

\paragraph{Bayesian CNN.}
In a Bayesian CNN, model parameters $\theta$ are considered as random variables, and the posterior distribution of $\theta$ is defined as:
\begin{equation}
p(\theta|\mathcal{D})=\frac{p(\mathcal{Y}|\mathcal{X},\theta)p(\theta)}{p(\mathcal{Y}|\mathcal{X})}   
\end{equation}
where $\mathcal{D}=\{\mathcal{X},\mathcal{Y}\}$ is the training data with inputs $\mathcal{X}$ and corresponding outputs $\mathcal{Y}$.

The inference in Bayesian CNN is conducted by averaging the prediction results over the posterior distribution $p(\theta|\mathcal{D})$, which is in practice intractable. Therefore, approximation methods have been developed, such as variational inference methods \cite{vatiation-inference} which approximate the true posterior distribution $p(\theta|\mathcal{D})$ with a tractable distribution $q(\theta)$ by minimizing their Kullback-Leibler (KL) divergence. 

In particular, Gal and Ghahramani \cite{dropout} constructed $q(\theta)$ with dropout that has been widely used as a stochastic regularization tool in deep learning. Such approximation has the benefit that a CNN model with dropout enabled can be easily cast as a Bayesian CNN without changing the model. Then, the predictive distribution of output $\boldsymbol{y}$ given a new input $\boldsymbol{x}$ can be obtained by Monte-Carlo integration as:
\begin{align}
\label{bnn}
\begin{split}
p(\boldsymbol{y}|\boldsymbol{x}) &= \int p(\boldsymbol{y}|\boldsymbol{x},\theta)q(\theta)\,d\theta\\
               &\approx \frac{1}{T}\sum_{t=1}^{T} p(\boldsymbol{y}|\boldsymbol{x},\theta_t), ~~\theta_t\sim q(\theta)
\end{split}
\end{align}
where T is the number of stochastic forward passes, and $\{ \theta_t\}_{t=1}^T \sim q(\theta)$ are the sampled parameters using dropout.

\section{Proposed method}

In this work, we propose an uncertainty-aware method for cross-domain object detection. We first compose a Bayesian CNN based framework for uncertainty estimation in object detection. Then uncertainty-aware pseudo-labels are generated by exploiting uncertainty in the target domain. Finally, the object detector is adapted to the target domain by joint feature alignment and self-training with the generated pseudo-labels. An overview of the proposed method is presented in Fig.~\ref{fig:model} with details introduced in the following.
 
\subsection{Uncertainty estimation in object detection}
\label{ssec_ueod}

In unsupervised domain adaptation, the uncertainty estimation is important since it measures the model's confidence about its predictions in the unseen target domain and hence could be used to guide the selection of ``confident'' pseudo-labels for re-training the model. Traditionally normalized output from standard CNNs is often erroneously used to interpret model uncertainty. Recent work \cite{dropout,Cai_2020_CVPR} have shown that a Bayesian CNN provides a more reliable way of uncertainty estimation by considering a distribution over model parameters. However, these work mainly considered uncertainty estimation in the classification/segmentation task. In object detection, model uncertainty comes from not only the classification branch but also the regression branch, both of which needs to be considered.

In this work, we provide a way of uncertainty estimation in object detection. The backbone model of object detector used in this work is based on Faster R-CNN \cite{ren2016faster}. The uncertainty is estimated by treating Faster R-CNN in a Bayesian framework (Bayesian Faster R-CNN). 
Given a candidate bounding box (or object instance) $\boldsymbol{b}^{rpn}$ from the Region Proposal Network (RPN), let us denote $\boldsymbol{g} \in \mathbb{R}^n$ to be a $n$-dimensional vector of detection scores from the classification branch and $\boldsymbol{l}=(u, v, w, h)$ to be the location output from the regression branch. Following Eq.~\ref{bnn}, the mean detection scores and location are obtained respectively by:

\begin{align}
\begin{split}
\label{un}
\boldsymbol{\bar g}  & \approx \frac{1}{T}\sum_{t=1}^{T} \boldsymbol{g}_t  =  \frac{1}{T}\sum_{t=1}^{T} C_{cls}(\boldsymbol{b},\boldsymbol{\theta}_t)  \\
\boldsymbol{\bar l} &\approx \frac{1}{T}\sum_{t=1}^{T} \boldsymbol{l}_t =  \frac{1}{T}\sum_{t=1}^{T} C_{reg}(\boldsymbol{b},\boldsymbol{\theta}_t) 
\end{split}
\end{align}
where $C_{cls}$ and $C_{reg}$ denotes the classification and the regression branches of the detector respectively.

Then a candidate pseudo-label $\boldsymbol{b}^{pl} = (\boldsymbol{\bar g}, \boldsymbol{\bar l})$ is obtained, with its uncertainty estimated by the predictive variance:

\begin{align}
\begin{split}
\label{uncer}
u  =&  ~u_{c}~+~u_{l} \\
 =& ~(\frac{1}{T}\sum_{t=1}^{T} {\boldsymbol{g}_t}^T\boldsymbol{g}_t -  {\boldsymbol{\bar g}}^T\boldsymbol{\bar g})~+~(\frac{1}{T}\sum_{t=1}^{T} {\boldsymbol{l}_t}^T\boldsymbol{l}_t - {\boldsymbol{\bar l}}^T\boldsymbol{\bar l} )
\end{split}
\end{align}
where $u_{c}$ and $u_{l}$ denotes the uncertainty for the classification and the regression respectively. $u$ is normalized to the range of [0,~1] for later usage.

\subsection{Uncertainty-aware pseudo-label selection} 
\label{ssec_pseudo-label}

Given a set of candidate pseudo-labels $\mathcal{Y}^{pl}=\{\boldsymbol{b}_i\}_{i=1}^{M}$, each containing a vector of detection scores and location of a bounding box, our goal is to select a subset $\mathcal{\hat Y}^{pl} \subset \mathcal{Y}^{pl}$ as pseudo-labels for self-training. 
At the core of our pseudo-label selection is the computation of selection scores $\mathcal{S}=\{s_i\}_{i=1}^{M}$ for all pseudo-labels. We got inspiration  from soft-NMS \cite{bodla2017soft} for computing the selection scores. The selection scores are initialized with the detection scores, and are updated in a recursive procedure. 
At each iteration, the bounding box $\boldsymbol{b}_m$ with maximum score $s_m$ is selected, and then the neighbouring bounding boxes are penalized whose overlap with $\boldsymbol{b}_m$ is larger than $\tau_1$. In \cite{bodla2017soft}, a Gaussian penalty function is used to significantly penalize overlap $iou(\cdot)$ that is close to one:

\begin{equation}
\label{lower}
    s_i = s_i \cdot e^{-\frac{iou(\boldsymbol{b}_m, \boldsymbol{b}_i)^2}{ \sigma}}
\end{equation}
where $\sigma$ is a constant hyper-parameter. 

Note that soft-NMS was originally proposed to generate more appropriate prediction for overlapping objects and has not been considered for pseudo-label selection. In this work, we adopt soft-NMS in our pseudo-label selection process and incorporate uncertainty $u_i$ as an adaptive attenuation factor to lower the selection score more rapidly for more uncertain samples, which is computed as:
\begin{equation}
\label{lower}
    s_i = s_i \cdot e^{-\frac{iou(\boldsymbol{b}_m, \boldsymbol{b}_i)^2}{ \sigma} \cdot e^{u_i}}
\end{equation}
We adopt the exponential form of $e^{u_i}$, so that the attenuation factor is larger than one. While there may also exists other ways of introducing uncertainty, we found current form effective in a preliminary study.

After each iteration, the bounding box $\boldsymbol{b}_m$ with its selection score $s_m$ is moved to the pseudo-label subset $\mathcal{\hat Y}^{pl}$, and the neighboring bounding boxes with selection scores less than a threshold $\tau_2$ will be discarded. The iteration would terminate when $\mathcal{Y}^{pl}$ is empty. 
Finally, we select at most $K$ bounding boxes from $\mathcal{\hat Y}^{pl}$ according to their selection scores as pseudo-labels. The procedure is shown in Algorithm \ref{argth}.

\begin{algorithm}
\caption{Uncertainty-aware pseudo-label selection}
\KwIn{
$\mathcal{Y}^{pl}=\{\boldsymbol{b_i}\}_{i=1}^{M}$, $\mathcal{S}=\{s_i\}_{i=1}^{M}$, $\tau_1$, $\tau_2$, $K$ \\
\quad \quad \quad$\mathcal{Y}^{pl}$: candidate pseudo-labels\\
\quad \quad \quad$\mathcal{S}$: detection scores\\
}
\KwOut{$\mathcal{\hat Y}^{pl}$}
$\mathcal{\hat Y}^{pl}\gets\{\}$ \\
\While{$\mathcal{Y}^{pl}\neq empty$}{
    $m\gets \argmax\limits_{i} s_i$ \\
    $\mathcal{\hat Y}^{pl} \gets \mathcal{\hat Y}^{pl}\cup (\boldsymbol{b_m}, s_m)$ \\
    $\mathcal{Y}^{pl} \gets \mathcal{Y}^{pl} - \boldsymbol{b_m} $  \\
    $\mathcal{S} \gets \mathcal{S} - s_m $  \\
    \For{$\boldsymbol{b_i} \in \mathcal{Y}^{pl} \land iou(\boldsymbol{b_m}, \boldsymbol{b_i}) \geq \tau_1$}{
        update $s_i$ according to Eq. \ref{lower} \\
        \If{$s_i < \tau_2$}{
             $\mathcal{Y}^{pl} \gets \mathcal{Y}^{pl} - \boldsymbol{b_i} $  \\
             $\mathcal{S} \gets \mathcal{S} - s_i $  \\
        }
    }
}
return maximum $K$ items from $\mathcal{\hat Y}^{pl}$
\label{argth}
\end{algorithm}

\subsection{Uncertainty-guided self-training}

Traditional self-training based methods rely on pseudo-labels generated based on the detection scores. This has one major drawback. Without the consideration of model uncertainty, noisy pseudo-labels may be selected and result in gradual drifts from self-training. Different from prior work, we explicitly consider uncertainty in pseudo-label selection.

With the procedure described in Section~\ref{ssec_pseudo-label}, a subset of pseudo-labeled samples. One straightforward way is to directly use them for self-training with equal weight. However, the difference and relative impact of the selected samples on training may be ignored. To make the self-training selectively focus on samples that the model is more confident on, we also impose sample-wise weight based on the estimated uncertainty. The weighted detection loss on the target data with pseudo-labels is defined as:
\begin{align}
\begin{split}
\label{wdet_loss}
     \mathcal{L}^T_{wdet}(\boldsymbol{x}^t, \mathcal{\hat Y}^{pl}) =  \sum_{\boldsymbol{b}^{pl}_i\in \mathcal{\hat Y}^{pl}}\ell_{det}(\boldsymbol{p}_i, \boldsymbol{b}^{pl}_i)\cdot w(\boldsymbol{\hat y}_i)
\end{split}
\end{align}
where $\ell_{det}$ denotes the instance-level detection loss, $\boldsymbol{p}_i$ and $\boldsymbol{b}^{pl}_i$ denote one prediction and the associated pseudo-label, $w(\boldsymbol{\hat y}_i)$ is the sample weight computed by one minus uncertainty score. Finally, the loss function for uncertainty-guided self-training is formulated as:
\begin{align}
\begin{split}
\label{ust_loss}
     &\mathcal{L}_{ust} = \mathcal{L}^S_{det}(\boldsymbol{x}^s, \mathcal{Y}^{gt}) + \lambda_1 \mathcal{L}^T_{wdet}(\boldsymbol{x}^t, \mathcal{\hat Y}^{pl}) 
\end{split}
\end{align}

\subsection{Uncertainty-guided feature alignment}
\label{ssec_ufa}

Feature alignment aims to learn domain-invariant features through adversarial learning, and has been developed as key techniques in the pioneer work \cite{chen2018domain} of cross-domain object detection. Specifically, image-level domain classifier $D_{img}$ and instance-level domain classifier $D_{ins}$ are constructed to align the feature distributions of two domains on the levels of image feature maps and object instances respectively. A consistency regularizer between the two level domain classifiers is also composed.
The feature alignment loss is formulated as:
\begin{align}
\begin{split}
\label{adaption_loss}
\mathcal{L}_{fa} &= \lambda_2 ( \mathcal{L}_{img} + \mathcal{L}_{ins} + \mathcal{L}_{cst})
\end{split}
\end{align}
where $\mathcal{L}_{img}$ and $\mathcal{L}_{ins}$ denote the image-level and instance-level adaptation losses respectively, $\mathcal{L}_{cst}$ denotes the consistency loss, $\lambda_2$ is a weight factor. For details of each loss component, please refer to \cite{chen2018domain}.

In this work, we further augment the feature alignment with our uncertainty-aware pseudo-labels. For instance-level feature alignment, it may be mistakenly conducted between a source domain object and a target domain background in \cite{chen2018domain}, since object instances are unknown in the target domain and the object proposals generated by RPN may not correspond to true objects. Therefore, we propose to utilize the pseudo-labels generated with our pseudo-label selection algorithm to mitigate the erroneous alignment.

By replacing RPN object proposals with pseudo-labeled object instances in this work, the instance-level adaptation loss is modified as:
\begin{align}
\begin{split}
    \mathcal{\hat{L}}_{ins}=  & -\sum_{\boldsymbol{b}^{gt} \in \mathcal{Y}^{gt}}\log\big(D_{ins}(\boldsymbol{b}^{gt} )\big)~-\\
&\sum_{\boldsymbol{b}^{pl} \in \mathcal{\hat Y}^{pl}}\log\big(1-D_{ins}(\boldsymbol{b}^{pl} )\big) 
\end{split}
\end{align}

The consistency loss is modified similarly for the target domain, and are denoted as $\mathcal{\hat L}_{cst}$.
Finally, the training loss for our uncertainty-guided feature alignment is:
\begin{align}
\begin{split}
\label{adaption_loss_new}
\mathcal{L}_{ufa} &= \lambda_2 ( \mathcal{L}_{img} + \mathcal{\hat{L}}_{ins} + \mathcal{\hat L}_{cst})
\end{split}
\end{align}

\subsection{Joint feature alignment and self-training}
In this work, we jointly consider feature alignment and self-training for domain adaptation. 
On the one hand, data discrepancy between different domains affects the prediction and hence the generation of pseudo-labels in the target domain, and feature alignment is an important step for generating reliable pseudo-labels for self-training. One the other hand, self-training is important for aligning the output distribution and learning discriminative features in the target domain. Therefore, we think feature alignment and self-training are complementary for successful domain adaptation and propose to combine the two parts together.
We construct a baseline model and a full model as described below, both considering joint feature alignment and self-training.

\paragraph{Our baseline model.} 
Our baseline model \textbf{$\mathcal{M}_1$} is trained with traditional self-training loss (Eq.~\ref{st_loss}) and feature alignment loss (Eq.~\ref{adaption_loss}). The baseline model is used mainly for examining how the proposed uncertainty-aware pseudo-labels can improve the performance. Note that the pseudo-labels in the self-training loss are selected based on detection scores. The loss fuction is formulated as:
\begin{align}
\begin{split}
\label{baseline}
     &\mathcal{L}_1=\mathcal{L}_{st} + \mathcal{L}_{fa}
\end{split}
\end{align}

\paragraph{Our full model.} Our full model \textbf{$\mathcal{M}_2$} is trained with uncertainty-guided self-training loss (Eq.~\ref{ust_loss}) and uncertainty-guided feature alignment loss (Eq.~\ref{adaption_loss_new}). Note that the pseudo-labels in the self-training loss are selected based on Algorithm~\ref{argth}. The loss function is formulated as:
\begin{align}
\begin{split}
\label{fullmodel}
    \mathcal{L}_2 = &\mathcal{L}_{ust} + \mathcal{L}_{ufa}
\end{split}
\end{align}

\begin{table*}[htb!]
    \centering
    \caption{Quantitative results on \textit{Citysacpes} $\rightarrow$  \textit{Foggy Cityscapes}.}
    \begin{tabular}{l|cccccccc|c}
    \toprule  
        Method & person & rider & car & truck & bus & train & motorbike & bicycle & mAP \\
     \midrule
    Faster R-CNN~\cite{ren2016faster} (Source) & 27.08 & 36.98 & 40.78 & 12.33 & 29.06 & 9.41 & 23.17 & 32.04 & 26.51\\
    \midrule  
    DA-Faster (CVPR'18)\cite{chen2018domain} & 29.60 & 40.40 & 43.40 & 19.70 & 38.30 & 28.50 & 23.70 & 32.70 & 32.00 \\
    Noisy Labeling (ICCV'19)\cite{khodabandeh2019robust} & 35.10 & 42.15 & 49.17 & \textbf{30.07} & 45.25 & 26.97 & 26.85 & 36.03 & 36.45 \\
    SWDA (CVPR'19) \cite{saito2019strong-weak} & 32.30 & 42.20 & 47.30 & 23.70 & 41.30 & 27.80 & 28.30 & 35.40 & 34.80 \\
    ICR-CCR (CVPR'20) \cite{xu2020categorical-regulaztion} & 32.90 & 43.80 & 49.20 & 27.20 & 45.10 &  36.40 & 30.30 & 34.60 & 37.40 \\
    GPA (CVPR'20)\cite{xu2020graph-induced} & 32.90 & 46.70 & 54.10 & 24.70 & 45.70 & 41.10 & 32.40 & 38.70 & 39.50 \\
    CT (ECCV'20)\cite{zhao2020collaborative} & 32.70 & 44.40 & 50.10 & 21.70 & 45.60 & 25.40 & 30.10 & 36.80 & 35.90 \\
    MeGA-CDA (CVPR'21) \cite{vs2021mega} & 37.70 & 49.00 & 52.40 & 25.40 & 49.20 & \textbf{46.90} & 34.50 & 39.00 & 41.80 \\
    \midrule
    Baseline model (ours) & 38.81 & 44.15 & 53.89 & 23.84 & 42.77 & 23.82 & 31.03 & 41.68 & 37.50 \\
    Full model (ours) & \textbf{42.19} & \textbf{51.19} & \textbf{58.84} & 25.64 & \textbf{49.99} & 33.96 & \textbf{37.03} & \textbf{45.24} & \textbf{43.01} \\
    \midrule
     Faster R-CNN (Oracle) & 40.63 & 47.05 & 62.50 & 33.12 & 50.43 & 39.44 & 32.57  & 42.43  & 43.52\\
    \bottomrule 
    \end{tabular}
    \label{tab:city-foggy}
\end{table*}

\section{Experiments}

\subsection{Datasets}
We conducted experiments on multiple public datasets to evaluate different scenarios of cross-domain object detection. In particular, we focus on object detection across different domains of driving video datasets which is of great importance in autonomous driving and remains as a challenging task:

\textbf{Cityscapes} \cite{cityscapes} is a real-world dataset for semantic urban scene understanding with images captured by a car-mounted camera. It has 2,975 images in the training set and 500 images in the validation set with pixel-level labels. 

\textbf{Foggy Cityscapes} \cite{foggy} is a foggy version of the Cityscapes dataset with same train/val split and annotations as those in the Cityscapes dataset.

\textbf{SIM10k} \cite{sim10k} is a synthetic dataset collected from a computer game and contains 10,000 images with 58,701 bounding boxes about cars.

\textbf{KITTI} \cite{geiger2013vision} is a real-world dataset in the field of autonomous driving, consisting of 7,481 annotated images.

\textbf{BDD100k} \cite{bdd100k} is a large driving video dataset with 100k annotated images. Following \cite{xu2020categorical-regulaztion}, we use the daytime subset of the dataset in our experiment, including 36,728 training and 5,258 validation images.

\subsection{Implementation details}
\paragraph{Network architecture.}
We adopt VGG16 \cite{simonyan2014very} as the backbone for Faster R-CNN. To formulate a Bayesian Faster R-CNN, we use the dropout layers that exist in the ROI based classifier, and utilize the network to generate pseudo-labels in testing time.
The networks of the image-level domain classifier and the instance-level domain classifier follow the setting in \cite{chen2018domain}. We additionally add dropout layers after the first convolutional layer in the image-level domain classifier and after the first two fully connected layers in the instance-level classifier.

\paragraph{Training details.}  
Each training batch is composed of one source domain image and one target domain image, which are resized with a length of 600 pixels. We use Adam optimizer with an initial learning rate of 0.002, a weight decay of 0.0005, and a momentum of 0.9. 
We first train the network with traditional feature alignment loss for 20 epochs, which is similar to \cite{chen2018domain}. Then we train our baseline model or full model for 5 epochs with model parameters initialized from the previous stage. 
We set $\lambda_1=0.001$ for the self-training loss, $\lambda_2=0.1$ for the feature alignment loss. We set $\tau_1 = 0.3$, $\tau_2 = 0.001$, $\sigma = 0.4$ and $K=20$ for our pseudo-label selection algorithm. The hyper-parameters are set with a preliminary empirical study and are fixed for later experiments.
We use Pytorch for implementation.

\subsection{Performance comparison}
We compare our baseline model (Eq.~\ref{baseline}) and full model (Eq.~\ref{fullmodel}) with Faster R-CNN \cite{ren2016faster} and recent state-of-the-art methods: 1) Adversarial feature alignment based methods including DA-Faster~\cite{chen2018domain}, SWDA~\cite{saito2019strong-weak}, ICR-CCR~\cite{xu2020categorical-regulaztion}, GPA~\cite{xu2020graph-induced}, and MeGA-CDA~\cite{vs2021mega}; 2) Self-training based methods including Noisy Labeling~\cite{khodabandeh2019robust}, and CT~\cite{zhao2020collaborative}.

\begin{table}[htb!]
    \centering
    \caption{Quantitative results on \textit{SIM10K} $\rightarrow$ \textit{Cityscapes}.}
    \begin{tabular}{l|c}
    \toprule
        Method & AP (car)\\
    \midrule
    Faster R-CNN~\cite{ren2016faster} (Source) & 34.60\\
    \midrule
    DA-Faster (CVPR'18)\cite{chen2018domain} & 41.90 \\
    Noisy Labeling (ICCV'19)\cite{khodabandeh2019robust} & 42.56 \\
    SWDA (CVPR'19) \cite{saito2019strong-weak} & 47.70 \\
    GPA (CVPR'20) \cite{xu2020graph-induced} & 47.60 \\
    CT (ECCV'20)\cite{zhao2020collaborative} & 44.51 \\
    MeGA-CDA (CVPR'21)\cite{vs2021mega} & 44.80\\
    \midrule
    Baseline model (ours) & 46.12\\
    Full model (ours) & \textbf{49.82}\\
    \midrule
    Faster R-CNN (Oracle) & 68.10 \\
    \bottomrule
    \end{tabular}
    \label{tab:sim10k-city}
\end{table}

\begin{table*}[htb!]
    \centering
    \caption{Quantitative results on \textit{Cityscapes} $\rightarrow$ \textit{BDD100k}.}
    \begin{tabular}{l|ccccccc|c}
    \toprule
    Method & person & rider & car & truck & bus &  motorbike & bicycle & mAP\\
    \midrule
    Faster R-CNN~\cite{ren2016faster} (Source) & 26.90 & 22.10 & 44.70 & 17.40 & 16.70 & 17.10 & 18.80 & 23.40 \\ 
    \midrule
    DA-Faster (CVPR'18)\cite{chen2018domain} & 29.40 & 26.50 &  44.60 & 14.30 & 16.80 & 15.80 & 20.60 & 24.00 \\
    SWDA (CVPR'19) \cite{saito2019strong-weak} & 30.20 & 29.50 & 45.70 & 15.20 & 18.40 & 17.10 & 21.20 & 25.30 \\
    ICR-CCR (CVPR'20) \cite{xu2020categorical-regulaztion} & 31.40 & 31.30 & 46.30 & \textbf{19.50} & \textbf{18.90} & 17.30 & 23.80 & 26.90 \\
    \midrule
    Baseline model (ours) & 35.55 & 30.66 & 51.69 & 17.49 & 15.16 & \textbf{19.27} & 26.57 & 28.06\\
    Full model (ours) & \textbf{37.28} & \textbf{32.86} & \textbf{55.79} & 18.97 & 15.36 & 17.59 & \textbf{26.98 }& \textbf{29.26}\\
    \midrule
     Faster R-CNN (Oracle) & 35.30 & 33.20 & 53.90 & 46.30 & 46.70 & 25.60 & 29.30 & 38.60 \\
    \bottomrule
    \end{tabular}
    \label{tab:city-bdd}
\end{table*}

Experiments are conducted on four different cross-domain scenarios. 
Following \cite{chen2018domain},  average precision (AP) of individual categories and mean average precision (mAP) with threshold of 0.5 are used as the evaluation metric.

\paragraph{Weather Adaption.} In this scenario, the performance across different weather conditions are evaluated. The Cityscapes dataset are used as the source domain and the Foggy Cityscapes dataset are used as the target domain. The results are shown in Table \ref{tab:city-foggy}. 
Our baseline model boosts the mAP by +5.5\% over DA-Faster~\cite{chen2018domain}, which validates the proposal of joint feature alignment and self-training. 
Meanwhile, our full model achieves the best performance of 43.01\% and largely outperforms our baseline model, verifying the effectiveness of the proposed method. Note that the oracle performance of training Faster R-CNN with target domain labels is only 43.52\% and there remains little room for further improvement.

\paragraph{Synthetic-to-Real Adaption.}
In this scenario, we evaluate the cross-domain performance from synthetic SIM10K dataset to real-world Cityscapes dataset. 
As shown in Table~\ref{tab:sim10k-city}, our method achieves the state-of-the-art performance with average precision of 49.82\%. Comparing with self-training based methods of Noisy Labeling~\cite{khodabandeh2019robust} and CT~\cite{zhao2020collaborative}, our method outperforms them by +7.26\% and +5.31\% respectively, demonstrating the importance of our uncertainty-guided self-training. Our full model also boosts the performance by +3.7\% over our baseline model. 

\paragraph{Cross-Camera Adaption.}
Here evaluate the cross-domain performance from one real-world KITTI dataset to another real-world Citiscapes dataset to examine the adaptation ability across different camera devices. Since the object categories in the two datasets are not the same, we evaluate the performance using average precision of car following \cite{chen2018domain}. The result are shown in Table \ref{tab:kitti-city}. Our method achieves state-of-the-art performance with average precision of 48.84\%.

\begin{table}[htb!]
    \centering
    \caption{Quantitave results on \textit{KITTI} $\rightarrow$ \textit{Cityscapes}.}
    \begin{tabular}{l|c}
    \toprule
        Method & AP (car) \\
    \midrule 
    Faster R-CNN~\cite{ren2016faster} (Source) & 37.60\\
    \midrule 
    DA-Faster (CVPR'18)\cite{chen2018domain} & 41.80 \\
    Noisy Labeling (ICCV'19)\cite{khodabandeh2019robust} & 42.98 \\
    GPA (CVPR'20)\cite{xu2020graph-induced} & 47.90 \\
    CT (ECCV'20)\cite{zhao2020collaborative} & 43.60 \\
    MeGA-CDA (CVPR'21)\cite{vs2021mega} & 43.00\\
    \midrule
    Baseline model (ours) & 43.56\\
    Full model (ours) & \textbf{48.84}\\
    \midrule
    Faster R-CNN (Oracle) & 68.10\\
    \bottomrule
    \end{tabular}
    \label{tab:kitti-city}
\end{table}

\paragraph{Scene and Scale Adaption.}
In this part, we examine the adaptation performance from a small-scale dataset (Cityscapes) to a large-scale dataset (BDD100k) recorded from different city scenes. As shown in Table \ref{tab:city-bdd}, our method achieves the highest mAP of 29.26\% and significantly outperforms existing methods. However, the absolute performance is still low. One possible reason might be that the domain shift between the two datasets is big and the multi-category adaptation task is more challenging. 

\begin{table}[htb!]
    \centering
    \caption{Ablation study on \textit{SIM10k} $\rightarrow$ \textit{Cityscapes}.}
    \begin{tabular}{l|c}
    \toprule
        Method &  AP \\
    \midrule 
    FA + ST & 46.12 \\
    UFA + ST &  48.40\\
    FA + UST & 49.03 \\
    UFA + UST & 49.82\\
    \bottomrule
    \end{tabular}
    \label{tab:ablation}
\end{table}

\begin{figure}[t]
    \centering
    \includegraphics[width=0.9\linewidth]{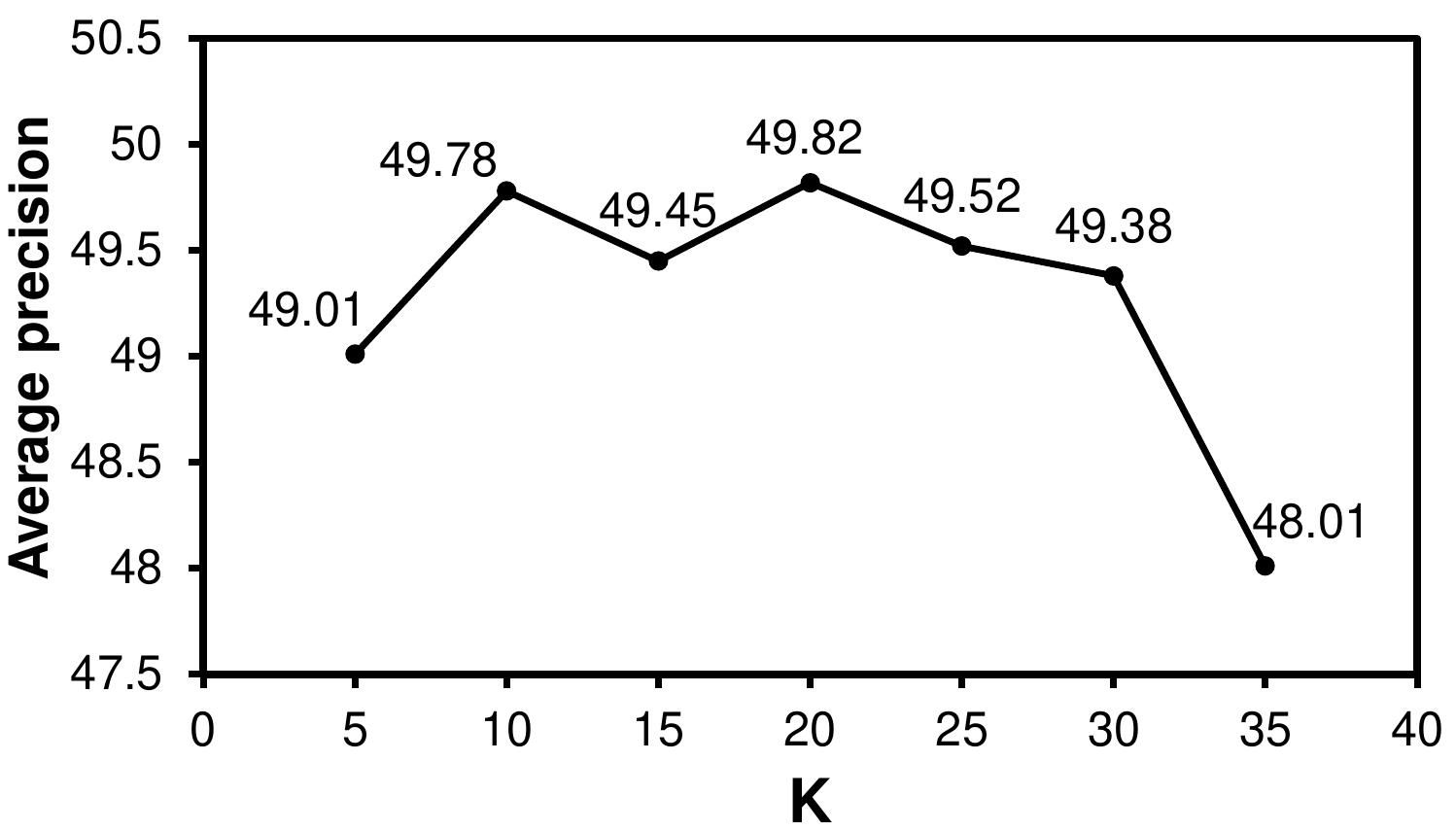}
    \caption{Performance variation with the different maximum number of pseudo-labels (K) on \textit{SIM10k} $\rightarrow$ \textit{Cityscapes}.}
    \label{fig:k_value}
\end{figure}

\begin{figure*}
    \centering
    \includegraphics[width=\linewidth]{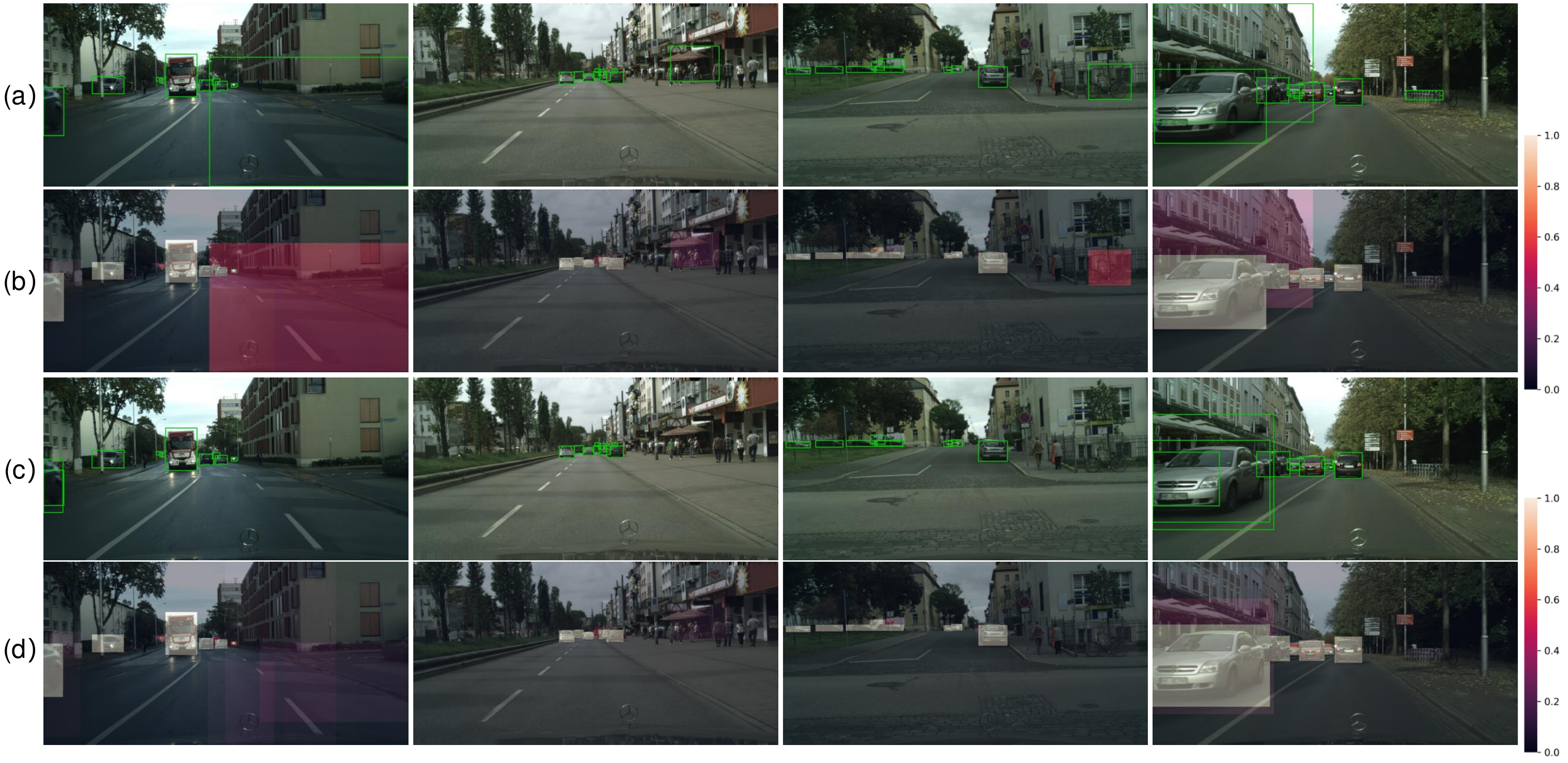}
    \caption{Examples of pseudo-labels for \textit{SIM10K} $\rightarrow$ \textit{Cityscapes} are obtained from traditional detection score-based pseudo-label selection method (a) and from our uncertainty-aware pseudo-label selection algorithm (c). Visualization of detection scores and selection scores of our algorithm for candidate objects are overlapped with original images and are shown in (b) and (d) respectively. The color maps for visualization in (b)(d) are shown at the rightmost.}
    \label{fig:sim10k_labels}
\end{figure*}

\subsection{Ablation study}
\label{ssec_ablation}

To investigate the effect of different modules in our method, we conduct an ablation study as follows:
\begin{itemize}
\vspace{-0.5em}
    \item{FA+ST:} Our baseline model of standard feature alignment and self-training with pseudo-labels selected based on detection scores.  (Eq.~\ref{st_loss} + Eq.~\ref{adaption_loss})
\vspace{-0.5em}
    \item{UFA+ST:} Uncertainty-guided feature alignment and standard self-training (Eq.~\ref{st_loss} + Eq.~\ref{adaption_loss_new}).
\vspace{-0.5em}
    \item{FA+UST:} Standard feature alignment and uncertainty-guided self-training (Eq.~\ref{ust_loss} + Eq.~\ref{adaption_loss}).
\vspace{-0.5em}
    \item{UFA+UST:} Our full model with uncertainty-guided feature alignment and self-training (Eq.~\ref{ust_loss} + Eq.~\ref{adaption_loss_new}).
\vspace{-0.5em}
\end{itemize}

The ablation study results on \textit{SIM10k} $\rightarrow$ \textit{Cityscapes} are shown in Table \ref{tab:ablation}. 
It can be seen that UFA+ST significantly outperforms the baseline model of FA+ST by +2.28\%, demonstrating the effectiveness of uncertainty-aware pseudo-labels for feature alignment. Similarly, performance improvement of +2.91\% is observed by comparing FA+UST with FA+ST, which indicates the importance of uncertainty-aware pseudo-labels for self-training. Lastly, our full model of UFA+UST achieves best performance and largely outperforms our baseline model by +3.7\%. These results indicate that model uncertainty is an important factor when considering joint feature alignment and self-training.

In our pseudo-label selection algorithm, we set a hyper-parameter of K to determine the maximum number of pseudo-labels to be used for uncertainty-guided self-training. Here we examine how different number of pseudo-labels affect the final performance. As shown in Fig.~\ref{fig:k_value}, the performance is relatively stable between $K=10$ and $K=30$, and then begins to drop quickly. This indicates that as K becomes larger over a certain threshold, the selection of noisy pseudo-labels becomes more possible and hence degrade the performance. 

\subsection{Qualitative analysis}
\label{ssec_visualization}

\begin{figure*}[t]
    \centering
    \includegraphics[width=0.95\linewidth]{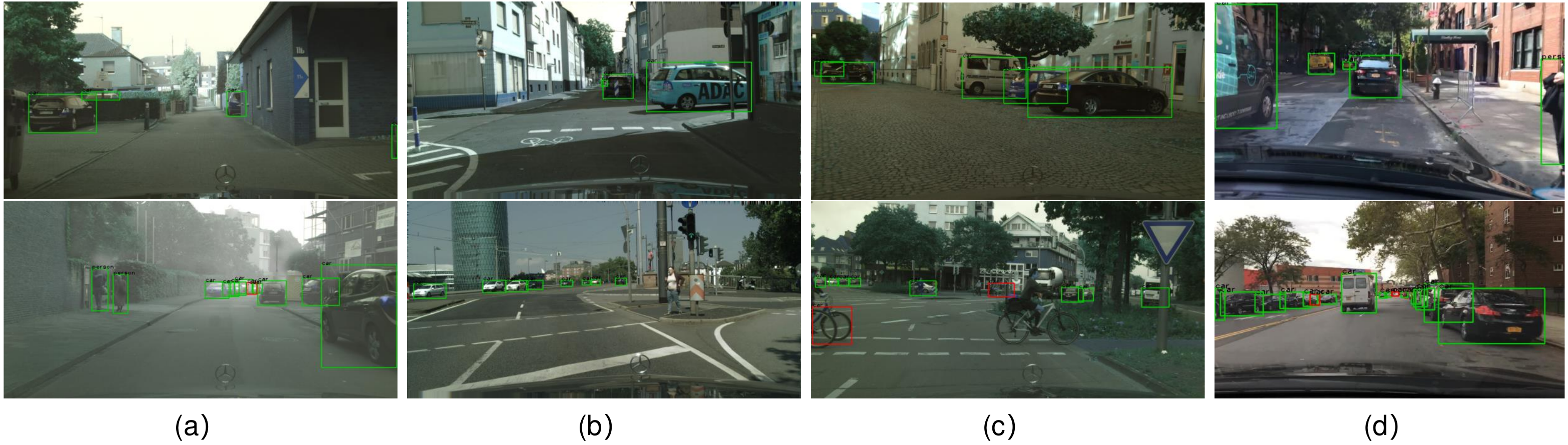}
    \caption{Examples of detection results for (a) \textit{Cityscapes} $\rightarrow$ \textit{Foggy Cityscapes}, (b) \textit{Sim10k} $\rightarrow$ \textit{Cityscapes}, (c) \textit{KTTI} $\rightarrow$ \textit{Cityscapes}, and (d) \textit{CityScapes} $\rightarrow$ \textit{BDD100k}. The color of a predicted bounding box is green if its intersection-over-union with the ground-truth bounding box is over 0.5, and red otherwise.}
    \label{fig:detection_results}
\end{figure*}

\paragraph{Visualization of pseudo-labels.} To analyze the behavior of our pseudo-label selection algorithm, we visualize the pseudo-labels as well as the detection scores and selection scores for \textit{SIM10K}$\rightarrow$ \textit{Cityscapes} in Fig.~\ref{fig:sim10k_labels}. For ease of viewing, we draw as many as 10 pseudo-labeled bounding boxes. It can be seen that detection score-based pseudo-label selection method (a) is prone to generate unreliable pseudo-labels, such as the background regions on the street or buildings. Our uncertainty-aware pseudo-label selection algorithm (c) can almost avoid such cases and generate more reliable pseudo-labels. Visualization of detection scores (b) and selection scores of our algorithm (d) helps explain the differences. Note that selection scores is computed based on detection scores by lowering the detection score of a candidate object with high uncertainty. Looking at the first example (first column of Fig.~\ref{fig:sim10k_labels}), while the detection score for the candidate object at the bottom-right of the image is high and may cause a false positive pseudo-label, the selection score is largely lowered and a false positive is avoided by considering uncertainty in our algorithm.

\paragraph{Qualitative detection results.} We show examples of detection results on different benchmarks in Fig.~\ref{fig:detection_results}. For ease of viewing, the number of predicted object bounding boxes to be drawn is kept the same as the number of ground-truth object bounding boxes. It can be seen that our method can correctly detect objects in different scenes. Failure cases are also observed in several challening cases. In (a) and (d), when the objects are overlapped and far away from the camera, the detection is not successful. In (c), the detector produces false prediction of cars for the regions that are partially occluded and have appearance of wheels.

\section{Conclusion}
In this paper, we tackle the problem of unsupervised cross-domain object detection, which is critical for deploying a pre-trained object detector to an unseen domain. We present an uncertainty-aware model adaptation method by constructing a Bayesian Faster R-CNN for uncertainty estimation in object detection. We compose a pseudo-label selection algorithm based on model uncertainty, and develop a method for joint feature alignment and self-training with uncertainty-aware pseudo-labels. Experiments on public benchmarks demonstrate that our method outperforms state-of-the-art methods on multiple adaptation scenarios.

{\small
\bibliographystyle{ieee_fullname}
\bibliography{arxiv}
}

\end{document}